\documentclass[letterpaper, 10 pt, conference]{ieeeconf}

\IEEEoverridecommandlockouts

\usepackage{graphicx}
\usepackage{bm}
\usepackage{amsmath}
\usepackage{amssymb}
\usepackage{cite}
\usepackage{url}
\usepackage[hidelinks]{hyperref}
\usepackage{booktabs}
\title{\LARGE \bf
Digital Twin-Driven VR Teleoperation with Multi-View Spatial Perception for Surgical Robots
}

\author{$^{*}$Chang Liu$^{1}$, $^{*}$Chenhao Yu$^{1}$, $^{*}$Honghao Zhao$^{2}$, Hao Ding$^{1}$, Haochen Wei$^{2}$, Adnan Munawar$^{1}$,\\ Mathias Unberath$^{1,2}$ and Peter Kazanzides$^{1,2}$
\thanks{$^{*}$These authors contributed equally to this work.}
\thanks{$^{1}$Laboratory for Computational Sensing and Robotics, Johns Hopkins University, Baltimore, MD, USA. Email: \texttt{cliu226@jhu.edu}}
\thanks{$^{2}$Department of Computer Science, Johns Hopkins University, Baltimore, MD, USA. Email: \texttt{pkaz@jhu.edu}}
}

\hypersetup{pdftitle={Digital Twin-Driven VR Teleoperation with Multi-View Spatial Perception for Surgical Robots},pdfauthor={Chang Liu, Chenhao Yu, Honghao Zhao, Hao Ding, Haochen Wei, Adnan Munawar, Mathias Unberath, Peter Kazanzides}}

\begin{document}

\maketitle

\thispagestyle{plain}
\pagestyle{plain}

\begin{abstract}

Current robot-assisted minimally-invasive surgery (RMIS) platforms provide a fixed console for the surgeon to view stereo endoscopic images and teleoperate instruments inside the patient. Several researchers have proposed the use of a head-mounted display (HMD) as a portable console, with video pass-through rendering of the endoscope images which, like the fixed console, restricts the operator to a single endoscopic viewpoint and limits depth perception. We present a digital twin-driven virtual reality (VR) teleoperation platform, where the digital twin is created from markerless perception of the surgical environment and streamed for display on the HMD. This overcomes the limitations of video pass-through by providing multi-view rendering and natural motion-parallax cues, enabling decoupling of the user's hand posture from strict instrument alignment. The system utilizes VR hand controllers to increase the teleoperation workspace and to improve the robustness and stability of instrument control compared to the hand tracking approach adopted by most prior systems.
A 15-participant user study on the da Vinci Research Kit (dVRK) shows that our VR platform significantly outperforms a state-of-the-art HoloLens~2 mixed reality baseline, reducing path length by 86\% and jerk by 95\%, while achieving depth perception confidence comparable to or exceeding the traditional console across all conditions.

\end{abstract}

\section{Introduction}
\label{sec:intro}
\begin{figure*}[!t]
    \centering
    \vspace{1.5mm}
    \includegraphics[width=0.85\textwidth]{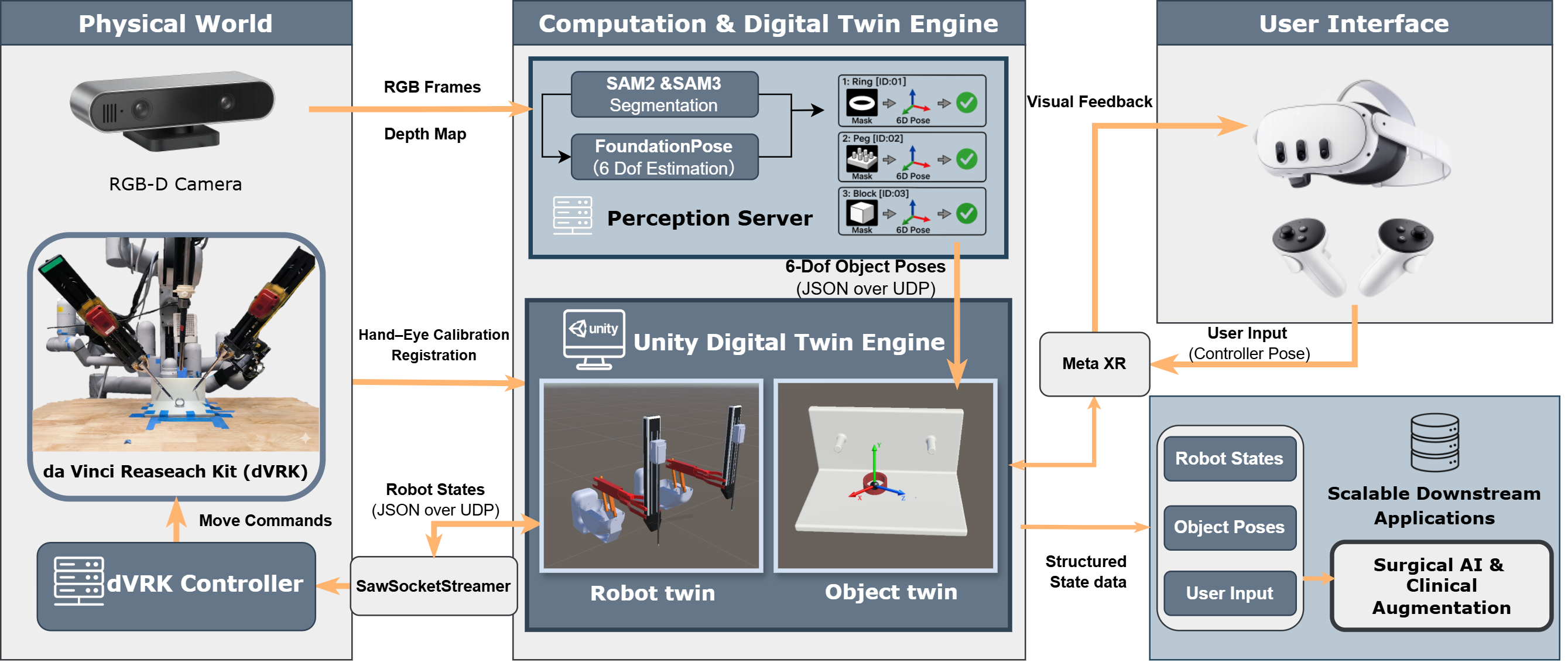}
\caption{System overview of the state-synchronized digital-twin teleoperation platform. Asynchronous streams drive the Unity digital twin for VR rendering and controller command generation, while lightweight structured states provide a foundation for scalable downstream applications.}
    \label{fig:system_overview}
\end{figure*}

Robot-Assisted Minimally Invasive Surgery (RMIS) provides surgeons with immersive 3D visualization and precise instrument control \cite{freschi2013davinci,kazanzides2014dvrk}, but current systems rely on bulky, stationary consoles that restrict operator mobility. To improve ergonomics and flexibility, recent research has explored ``console-free'' teleoperation using head-mounted displays (HMDs) \cite{magnani2026consolefree,qian2019dvrkxr,dardona2019remotepresence,rhee2024unconstrained}.

A fundamental challenge of surgical teleoperation is that the endoscopic viewpoint offers limited spatial cues: it provides a single, relatively narrow perspective whose repositioning requires deliberate effort, depriving the operator of the continuous depth and parallax information needed for precise manipulation \cite{almeida2020teleoptransparency,warburton2023mtp}. Moreover, the endoscope forces the surgeon to operate from a video feed rather than looking directly at their own hands, fundamentally disrupting natural
hand--eye coordination \cite{lanfranco2004robotic,guthart2000intuitive}. The conventional da Vinci console overcomes this by kinematically aligning the master manipulators with the endoscopic view \cite{guthart2000intuitive}, creating the intuitive sensation that ``my hands are the instruments''. This tight hand--instrument correspondence gives the operator reliable proprioceptive cues that compensate for the limited visual depth \cite{freschi2013davinci,guthart2000intuitive}, but it ties the surgeon to a bulky, stationary console.

Console-free MR systems often attempt to preserve some form of hand-instrument coupling to compensate for the constrained visual feedback of video pass-through \cite{magnani2026consolefree,ChenISMR2023,ai2024mrteleop,borgioli2025glove}. However, the practicality of these systems is further hindered by their reliance on bare-hand gesture tracking. Mid-air hand motions are inherently susceptible to tremors, tracking loss, and occlusion \cite{ChenISMR2023,rhee2024unconstrained,khundam2021vrcontroller}. Furthermore, the absence of physical controllers makes it difficult to implement a reliable \emph{clutch} mechanism \cite{zhou2016gesture,fu2021mobile}. Taken together, these visual and mechanical limitations have made truly practical console-free teleoperation elusive.

To address these challenges, we propose a Digital Twin-Driven Virtual Reality (VR) teleoperation platform for the da Vinci Research Kit (dVRK). Instead of streaming endoscopic video, our system uses real-time markerless perception to construct a digital twin of the surgical scene inside a Meta Quest 3S headset. This architecture natively supports multi-view rendering and natural motion-parallax cues~\cite{rogers1979motionparallax}, restoring the rich spatial understanding missing in traditional video pass-through.

Building upon this platform, we investigate a novel teleoperation control paradigm that safely decouples the user's hand posture from strict instrument alignment. Utilizing a clutch-anchored relative mapping with physically grounded 6-DoF controllers, operators can manipulate instruments intuitively from comfortable, unconstrained poses. Our findings highlight that enhanced spatial understanding is crucial for users, as rich visual spatial cues can successfully offset the traditional reliance on physical hand-instrument correspondence.

Furthermore, the state-synchronized architecture naturally logs structured kinematic and spatial states, establishing a scalable pipeline for multi-view training data generation and potential integration as an auxiliary 3D spatial view in conventional clinical workflows.

The main contributions of this work are:
\begin{itemize}
    \item A digital twin-driven VR teleoperation framework that replaces video streaming with state synchronization, enabling unconstrained multi-view rendering for enhanced depth and spatial perception.
    \item A clutch-anchored teleoperation paradigm that leverages this enhanced spatial awareness to safely decouple user hand posture from instrument alignment via stable 6-DoF controllers.
    \item A 15-participant user study demonstrating that VR significantly outperforms a state-of-the-art MR baseline in trajectory quality, with spatial judgment comparable to the traditional console.
\end{itemize}

\section{Related Work}
\label{sec:related_work}

\subsection{Extended Reality (XR) and Surgical Teleoperation Interfaces}

Early efforts to decouple the operator from the surgical console explored HMDs for remote endoscope camera control \cite{dardona2019remotepresence}. Building on MR frameworks such as dVRK-XR \cite{qian2019dvrkxr}, several console-free teleoperation systems have since been proposed. Chen et al.~\cite{ChenISMR2023} used HoloLens~2 hand tracking with relative position and orientation control but found the approach unintuitive when the user's hand orientation diverged from the instrument's. Ai et al.~\cite{ai2024mrteleop} extended this work by adding endoscopic video visualization and head-tracked camera control, but relied on absolute orientation mapping, which was limited by hand-tracking occlusion and exhibited notable latency. Rhee et al.~\cite{rhee2024unconstrained} developed a lightweight gesture interface using MediaPipe, though the system was validated only in simulation. Borgioli et al.~\cite{borgioli2025glove} achieved MTM-comparable performance using sensory gloves and external trackers, but the system required multiple hardware components and exhibited approximately 200\,ms latency.

A recurring theme across these systems is the need to maintain a visual or physical correspondence between the user's hands and the instruments. When porting teleoperation to HMDs, the limited spatial feedback of video pass-through makes some form of coupling essential for the user's spatial orientation. Magnani et al.~\cite{magnani2026consolefree} explicitly addressed this by rendering virtual instrument shafts aligned with the physical tools for HoloLens~2-based teleoperation. We adopt their system as our MR baseline (Sec.~\ref{sec:experiments}).

While these systems improve portability, their reliance on video-based rendering restricts the operator to a single viewpoint \cite{almeida2020teleoptransparency,warburton2023mtp}, and the lack of dedicated handheld controllers hinders reliable clutch mechanisms \cite{zhou2016gesture,fu2021mobile}. Our platform addresses both limitations by rendering a digital twin with flexible viewpoints and utilizing 6-DoF VR controllers for robust input.

\subsection{Surgical Scene Perception and Digital Twins}

A functional digital twin requires continuous, accurate mapping of the physical environment. Conventional AR overlays in surgery generally depend on physical fiducials, such as ArUco markers or optical tracking spheres. The emergence of Vision Foundation Models (VFMs) provides a robust alternative for markerless tracking. The Segment Anything Model (SAM) \cite{kirillov2023sam} offers zero-shot semantic segmentation, and FoundationPose \cite{wen2024foundationpose} tracks 6D object poses without object-specific retraining.

Recent work has shown that digital-twin representations derived from foundation models can support offline surgical task automation and planning~\cite{ding2025towards}. Digital twins have also been explored for surgical robotics training with contextual assistance~\cite{hagmann2021dtassistance} and VR-based RMIS simulation~\cite{cai2023vrdtrmis}. In parallel, synchronized multi-modal dVRK data frameworks have shown that temporally aligned vision, kinematics, and auxiliary sensing can support downstream learning and real-time inference~\cite{zhou2026surgsync}. However, bringing these perception-driven scene representations into a real-time robotic control loop for live teleoperation remains computationally demanding. Our framework integrates these VFMs directly into the live teleoperation pipeline, utilizing real-time 3D object tracking to replace traditional video pass-through and provide the rich spatial feedback that underpins our approach to console-free control.

\section{System Overview}
\label{sec:system_overview}

Figure~\ref{fig:system_overview} summarizes the end-to-end pipeline across the physical layer, computation layer, and user interface. Our platform consists of three asynchronous flows:
(i) Robot state stream: the dVRK publishes high-rate robot joint states as JSON packets over UDP via \texttt{sawSocketStreamer}\footnote{\url{https://github.com/jhu-saw/sawSocketStreamer}}, which are consumed by Unity to update the robot digital twin in real time;
(ii) Object state stream: an external RGB-D camera observes the workspace and sends images to a perception server, where segmentation and 6-DoF pose tracking produce rigid object transforms that are serialized and streamed to Unity to update the object digital twin;
(iii) Command stream: Meta Quest~3S controller poses and button inputs are mapped in Unity to dVRK-compatible motion commands and transmitted back to the robot through the same UDP socket interface, closing the teleoperation loop.

A key design principle is to decouple \emph{rendering} from \emph{state transport}: Unity receives state packets asynchronously and renders a state-consistent scene at the headset refresh rate, while the robot and perception streams can operate at their own update rates. The same structured state streams can also be recorded for
lightweight logging and offline replay.

Accurate real--virtual alignment is required for a faithful digital twin. Because the kinematics of the passive Setup Joints (SUJ) are not accurate enough for precise registration, relying on them causes drift between the virtual and physical robots. To resolve this, we bypass the SUJ entirely and establish a shared, camera-centric coordinate system via hand-eye calibration.

\subsection{dVRK Digital Twin}
\label{sec:robot_twin}

The robot digital twin is implemented in Unity as a kinematic model consistent with the dVRK geometric and kinematic specifications (e.g., link dimensions and joint limits) and includes the Patient Side Manipulator (PSM)'s Remote Center of Motion (RCM) constraint. To synchronize the full arm configuration, we adopt a joint-state-driven update at runtime: upon receiving a robot state packet, Unity parses the measured joint angles $\bm{q}_{meas}$ and applies them to the corresponding joints in the Unity kinematic chain, then evaluates forward kinematics to update the pose of every link. This whole-chain synchronization allows the operator to visually reason about arm configuration, joint-limit proximity, and potential collision risks in the reconstructed workspace.

\subsection{Object State Twin}
\label{sec:object_twin}

\begin{figure}[!t]
    \centering
    \includegraphics[width=0.9\linewidth]{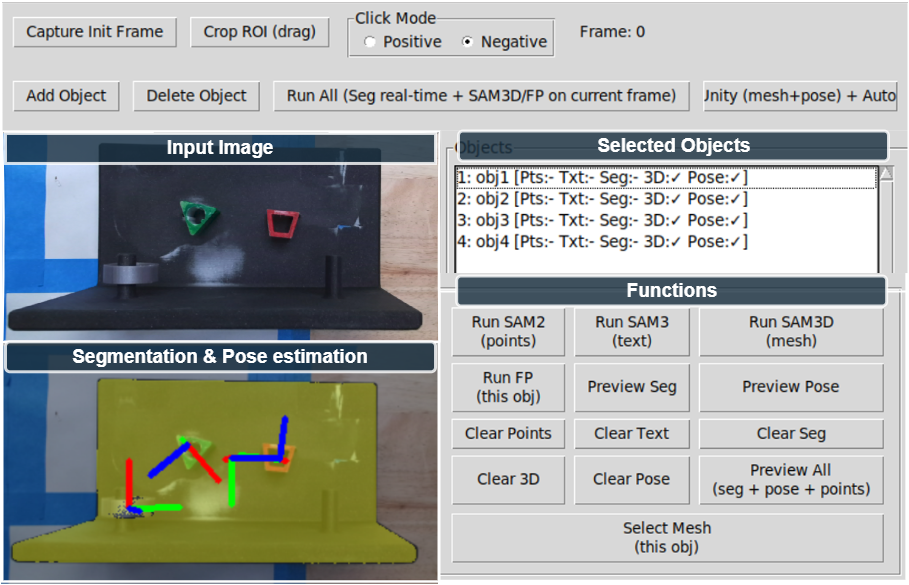}
    \caption{Perception GUI for managing multiple object digital twins. }
    \label{fig:dt_gui}
\end{figure}

In addition to the robot kinematic twin, the system maintains an object state twin. An external RGB-D camera provides frames to a perception server, which estimates the 6-DoF poses of manipulated items and streams lightweight state packets (JSON over UDP) to Unity. To manage the perception pipeline, we developed a custom graphical user interface (Fig.~\ref{fig:dt_gui}) that supports multi-object tracking, confidence monitoring, and manual re-initialization. The pipeline consists of three main components:

\subsubsection{Segmentation and Tracking (SAM 2 \& SAM 3)}
We leverage SAM~2~\cite{ravi2024sam2} to obtain object-centric instance masks to isolate targets from background clutter. At initialization, an operator provides a minimal prompt (e.g., point or box) to identify the object in the camera view. The segmentation module then propagates the mask over time to maintain a consistent instance track, producing a per-frame binary mask for each object. These masks stabilize object identity during motion and partial occlusions and serve as a spatial prior for downstream pose estimation.

Optionally, SAM~3 semantic segmentation provides class-level labels (e.g., tool, ring, block) to improve bookkeeping when multiple similar objects are present.

\subsubsection{6D Pose Estimation and State Streaming (FoundationPose)}
Given the instance mask and the current RGB-D frame, we estimate the 6D pose of each object using FoundationPose \cite{wen2024foundationpose}. The method leverages an object geometry prior (CAD mesh or 3D model) and performs sampling/refinement to recover a rigid transform without object-specific retraining. For each tracked object $o$, the perception server outputs
\begin{equation}
{}^{C}\mathbf{T}_{o}(t) \in SE(3),
\end{equation}
expressed in the RGB-D camera frame $\{C\}$.

Each state packet includes an object ID, timestamp, translation $\mathbf{p}\in\mathbb{R}^3$, rotation, and optional semantic metadata (e.g., class label $c_o$ and confidence scores). Unity receives packets asynchronously and updates the corresponding virtual object's transform. With the camera-centric registration described in Sec.~\ref{sec:handeye_registration}, object updates are direct:
\begin{equation}
{}^{W}\mathbf{T}_{o}(t) = {}^{C}\mathbf{T}_{o}(t),
\end{equation}
requiring no additional object-side spatial calibration beyond establishing the shared camera frame.

\subsubsection{Multi-Object Management and Robust Rendering}
Since the perception update rate is typically lower than the VR rendering rate, Unity buffers the incoming timestamped states. We apply a first-order low-pass filter for translation and quaternion SLERP for rotation to maintain smooth visualization. The system holds the last valid pose during temporary tracking dropouts and alerts the operator via the GUI (Fig.~\ref{fig:dt_gui}) if manual re-initialization is needed. When multiple objects are tracked, a DT manager assigns persistent IDs and can use SAM~3 semantic labels to reduce ID switches among similar instances.

\subsection{Hand--Eye Calibration and Registration}
\label{sec:handeye_registration}

\begin{figure}[!t]
    \centering
    \includegraphics[width=0.9\linewidth]{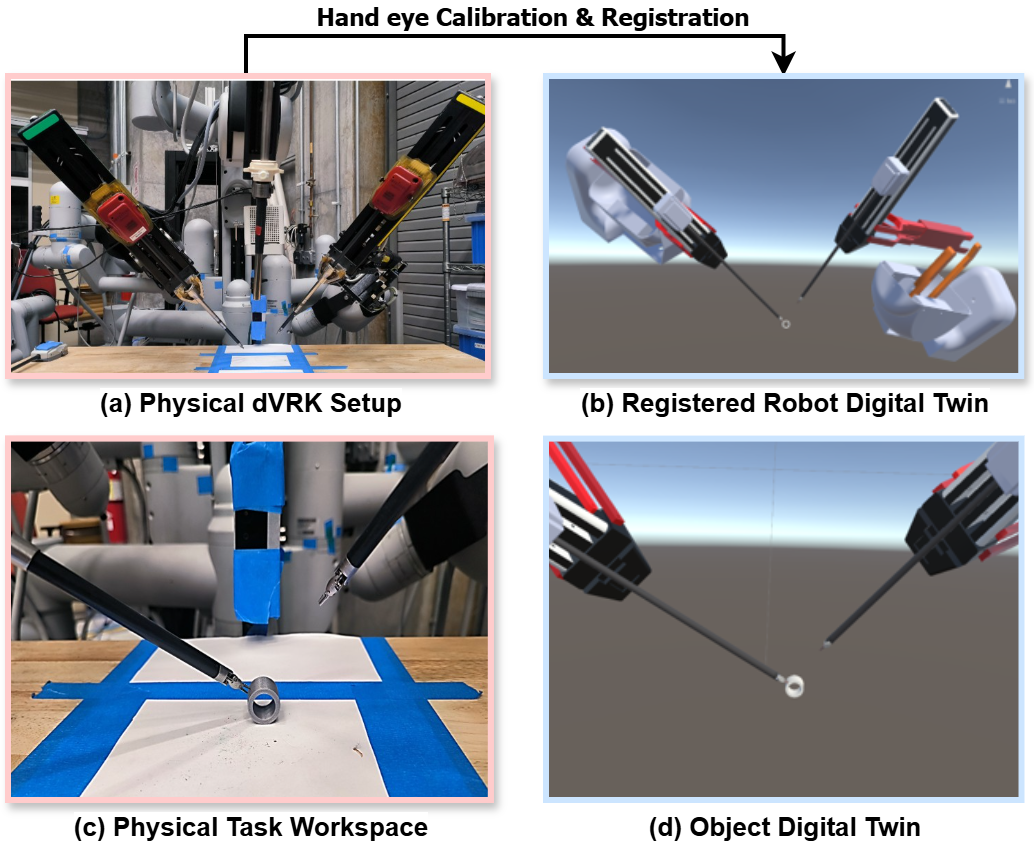}
    \caption{Registration of the digital twin. The physical dVRK (a) and workspace (c) are aligned with their virtual counterparts in Unity (b, d) using a unified camera-centric coordinate frame.}
    \label{fig:registration_qualitative}
\end{figure}

To align the physical robot, objects, and their digital counterparts, we perform hand--eye calibration and use the result to redefine the dVRK kinematic base frame, bypassing the inaccurate Setup Joint (SUJ) kinematics so that perception and control share a unified camera-centric coordinate system (Fig.~\ref{fig:registration_qualitative}).

\subsubsection{Hand--Eye Calibration }
Let $\{C\}$, $\{B\}$, $\{Tip\}$, and $\{M\}$ denote the coordinate frames of the tracking (RGB-D) camera, the PSM kinematic reference frame (RCM), the tool tip, and the calibration marker (attached to the tool shaft), respectively. The kinematic chain is:
\begin{equation}
{}^{C}\mathbf{T}_{M} = {}^{C}\mathbf{T}_{B} \cdot {}^{B}\mathbf{T}_{Tip} \cdot {}^{T}\mathbf{T}_{M}.
\label{eq:kinematic_chain}
\end{equation}
Here ${}^{C}\mathbf{T}_{M}$ is the marker pose estimated from the camera (PnP), and ${}^{B}\mathbf{T}_{Tip}$ is the tool-tip pose from robot forward kinematics. The marker is rigidly attached to the tool, thus ${}^{Tip}\mathbf{T}_{M}$ is constant but unknown. We therefore estimate ${}^{C}\mathbf{T}_{B}$ using a standard hand--eye calibration over $N$ synchronized pose samples, jointly accounting for the constant tool-to-marker offset.

\subsubsection{Robot Base Registration}
In the standard dVRK control architecture, the robot's Cartesian positions are typically expressed relative to the Endoscopic Camera Manipulator (ECM) tip or the world frame via the Setup Joints (SUJ). Since the SUJ  kinematics are not sufficiently accurate for precise registration, we bypass them entirely and reconfigure the dVRK console by defining the RGB-D camera coordinate system $\{C\}$ as the new global reference. Using the calibrated transform ${}^{C}\mathbf{T}_{B}$ obtained above, we modify the robot's base frame definition in the controller configuration files. Consequently, the kinematic stream output by the dVRK controller,${}^{C}\mathbf{T}_{Tip}$, is directly expressed in the camera frame:
\begin{equation}
{}^{C}\mathbf{T}_{Tip} = {}^{C}\mathbf{T}_{B} \cdot \text{FK}(\bm{q})
\end{equation}
where $\text{FK}(\bm{q})$ represents the forward kinematics from the base to the tip.

In the Unity digital twin, we define the World Origin as the Camera Origin (i.e., $\{W\} \equiv \{C\}$). We import the calibration result ${}^{C}\mathbf{T}_{B}$ to set the pose of the virtual PSM base game object relative to the Unity camera. As seen in Fig.~\ref{fig:registration_qualitative}(b), this ensures the virtual robot is anchored correctly in the visual space.

\subsubsection{Object Digital Twin Registration}
For the object digital twin described in Sec.~\ref{sec:object_twin}, registration is implicit and seamless due to our camera-centric architecture. The perception pipeline (FoundationPose) estimates the object pose ${}^{C}\mathbf{T}_{O}$ directly in the camera coordinate frame.
Since the Unity world origin coincides with the camera origin, no additional spatial calibration is required for the objects.  This ensures that the virtual objects appear in the exact physical locations relative to the robot, as verified by the alignment between the virtual tool-tip and the ring in Fig.~\ref{fig:registration_qualitative}(d).

\subsection{VR Teleoperation Interface}
\begin{figure}[!t]
    \centering
    \includegraphics[width=0.9\linewidth]{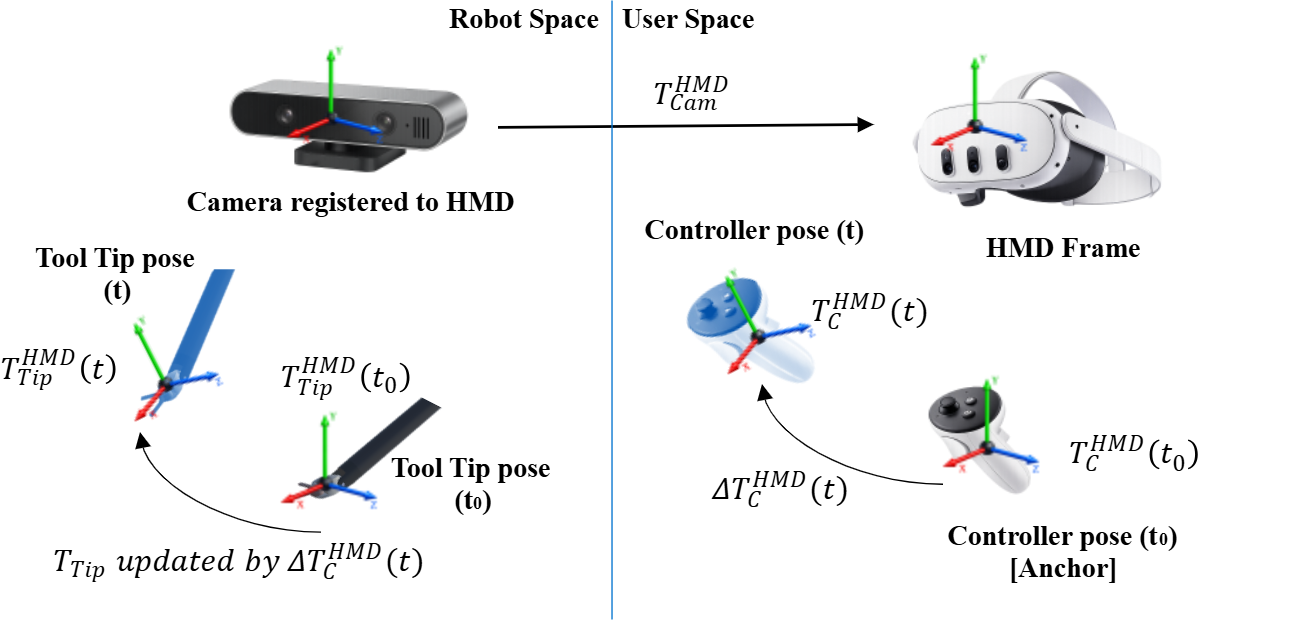}
\caption{Clutch-anchored motion mapping. Controller displacements relative to the engagement anchor are mapped to the robot tip within the unified HMD coordinate frame.}
    \label{fig:controller_mapping}
\end{figure}
\subsubsection{Controller-to-Robot Motion Mapping}
As illustrated in Fig.~\ref{fig:controller_mapping}, we map the handheld controller motion to the PSM tip using a \emph{clutch-anchored relative pose mapping} defined in the headset tracking (HMD) coordinate frame. When the clutch is engaged at time $t_0$, the system records the initial controller pose $T^{HMD}_{C}(t_0)$ and the initial robot tip pose $T^{HMD}_{Tip}(t_0)$. During clutch engagement, the controller motion is interpreted relative to this anchor:
\begin{align}
\Delta T^{HMD}_{C}(t)
&=
\left(T^{HMD}_{C}(t_0)\right)^{-1}
T^{HMD}_{C}(t)
\\
&=
\begin{bmatrix}
\Delta R^{HMD}_{C}(t) & \Delta p^{HMD}_{C}(t) \\
\mathbf{0} & 1
\end{bmatrix}.
\end{align}
This yields a translational displacement $\Delta p^{HMD}_{C}(t)$ and a rotational displacement $\Delta R^{HMD}_{C}(t)$ with respect to the clutch start.

The target tip pose is then computed by composing the anchored robot pose with these relative displacements. Specifically, translation is applied with a scale factor $s$:
\begin{align}
p^{HMD}_{Tip}(t) = p^{HMD}_{Tip}(t_0) + s\,\Delta p^{HMD}_{C}(t),
\end{align}
and rotation is applied by left-composition:
\begin{align}
R^{HMD}_{Tip}(t) = \Delta R^{HMD}_{C}(t)\, R^{HMD}_{Tip}(t_0).
\end{align}

This clutch-anchored formulation provides three key advantages. First, computing the target pose relative to a fixed anchor ($t_0$) rather than via frame-to-frame integration reduces sensitivity to tracking jitter, yielding smoother trajectories. Second, because teleoperation depends only on relative motion, no absolute spatial alignment between controller and robot workspace is required, enabling seamless re-anchoring. Third, and most importantly, our formulation clutches \emph{both} translation and rotation simultaneously. The conventional da Vinci paradigm maintains absolute rotational mapping even during clutching to preserve the ``hands are the instruments'' mental model.  In contrast, operators under the VR scene can easily comprehend the instrument's state visually. This allows users to freely reposition and reorient their hands to the most ergonomic posture during clutching, fundamentally breaking the strict physical hand-instrument correspondence without sacrificing control intuitiveness.

\begin{figure}[!t]
    \centering
    \includegraphics[width=0.8\linewidth]{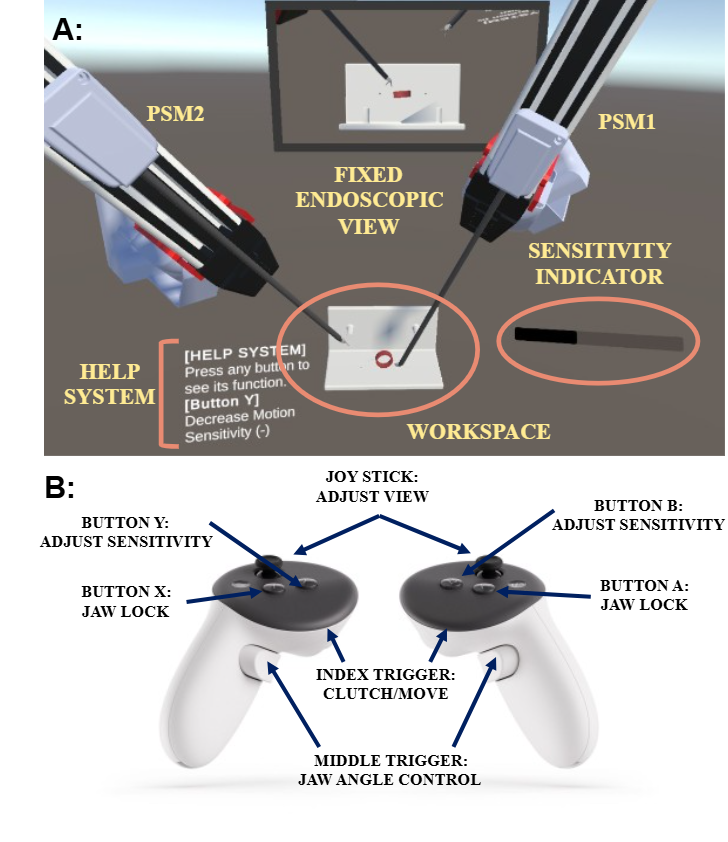}
    \caption{Teleoperation interface. (A) The VR UI provides multi-view visual feedback and real-time system status. (B) Mapping of Meta Quest 3S controller inputs for clutching, jaw control, and view adjustments.}
    \label{fig:ui_controls}
\end{figure}

\subsubsection{User Interface and Teleoperation Controls}
As shown in Fig.~\ref{fig:ui_controls}, we provide a lightweight VR UI and a compact set of controller inputs for safe and ergonomic teleoperation. Users can switch among discrete motion scale levels using buttons Y/B, and the current level is displayed in the UI (Fig.~\ref{fig:ui_controls}A), enabling seamless transitions between coarse repositioning and fine alignment.

\subsubsection{Grasp Feedback and Jaw Lock}
Reliable confirmation of grasp is critical in a fully digital interface. In our implementation, the gripper jaw is controlled by the middle trigger as an analog input. Let $u(t)\in[0,1]$ denote the middle-trigger press depth (0: released, 1: fully pressed). The desired jaw angle command is computed by linearly mapping $u(t)$ from a calibrated open/close range:
\begin{align}
q_{\mathrm{des}}(t) &= \Big(\, \theta_{\max} - (\theta_{\max}-\theta_{\min})\,u(t)\,\Big)\cdot \frac{\pi}{180},
\label{eq:jaw_map}
\end{align}
where $\theta_{\min}=-10^\circ$ and $\theta_{\max}=60^\circ$ are the jaw-angle limits used in the runtime mapping.

To confirm grasp, we detect jaw stall at a non-fully-closed angle (indicating object contact) and trigger a haptic vibration pulse. A jaw-lock mode (buttons X/A, Fig.~\ref{fig:ui_controls}B) latches the current jaw angle to prevent accidental reopening during transport.

\subsubsection{Multi-View Rendering and State-Driven Replay}
Unlike video-based teleoperation that restricts the operator to a single endoscopic viewpoint, our digital-twin VR interface provides a native workspace view as the primary visualization. In addition, we render an extra fixed endoscopic view (Fig.~\ref{fig:ui_controls}A) to emulate the conventional surgical camera perspective. Presenting both views simultaneously allows operators to cross-check depth and alignment from complementary perspectives, significantly improving spatial awareness and reducing occlusion-related ambiguity during manipulation.

\begin{figure}[!t]
    \centering

    \includegraphics[width=0.8\columnwidth]{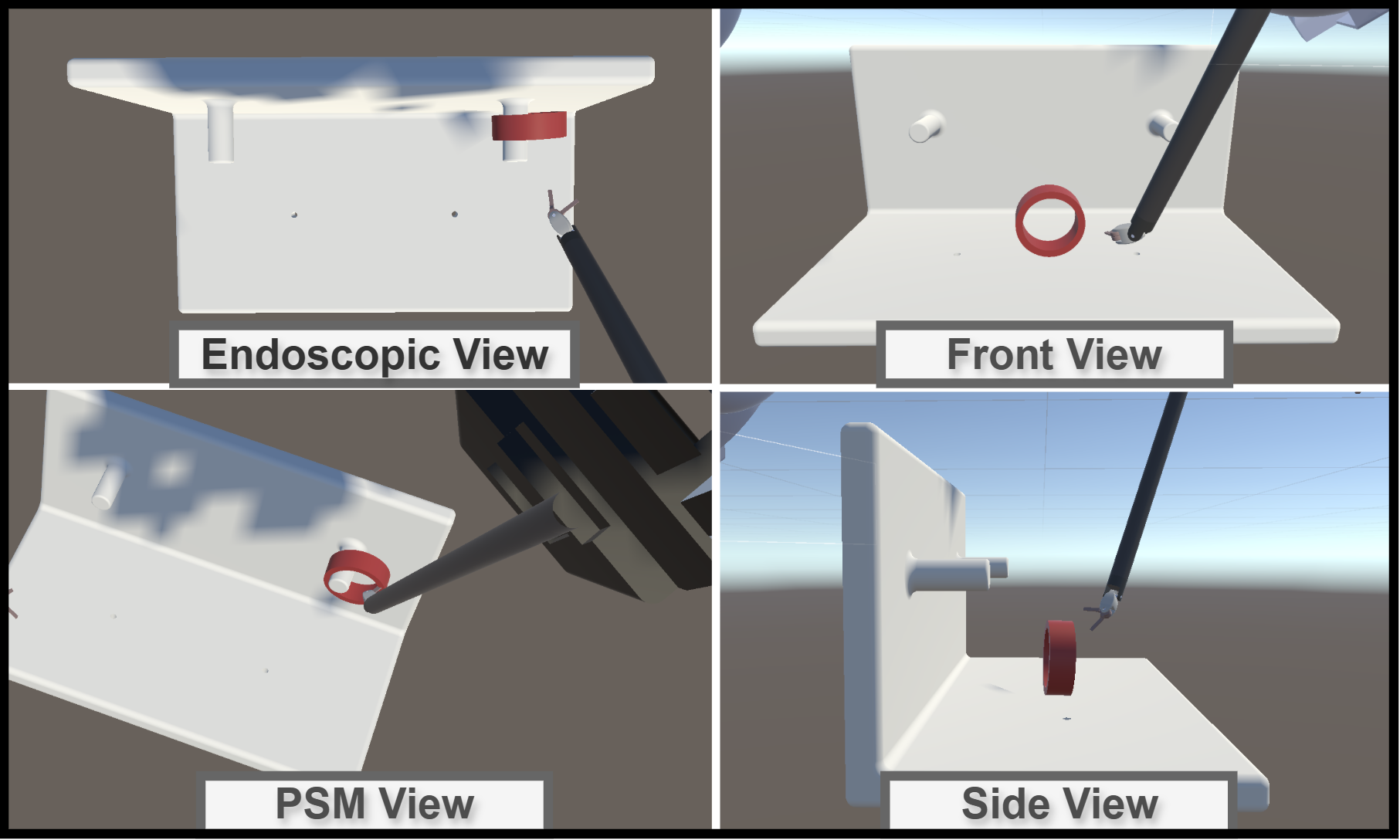}
    \caption{Offline multi-view replay. Four sample views are presented.}
    \label{fig:multi_view}
\end{figure}

Beyond real-time visualization, this state-synchronized architecture inherently unlocks flexible offline functionalities. Because the digital twin is driven by lightweight kinematic and spatial state streams rather than high-bandwidth video, the system supports structured data logging. By parsing these saved state sequences, recorded physical demonstrations can be seamlessly replayed offline within the Unity environment. During this replay phase, multiple virtual cameras can be deployed to synchronously render the task from arbitrary, unconstrained perspectives  (Fig.~\ref{fig:multi_view}). This demonstrates the functional versatility of the platform.

\section{EXPERIMENTS}
\label{sec:experiments}
\begin{figure*}[!t]
    \centering

    \includegraphics[width=0.8\linewidth, trim=1cm 0.7cm 1cm 0.2cm, clip]{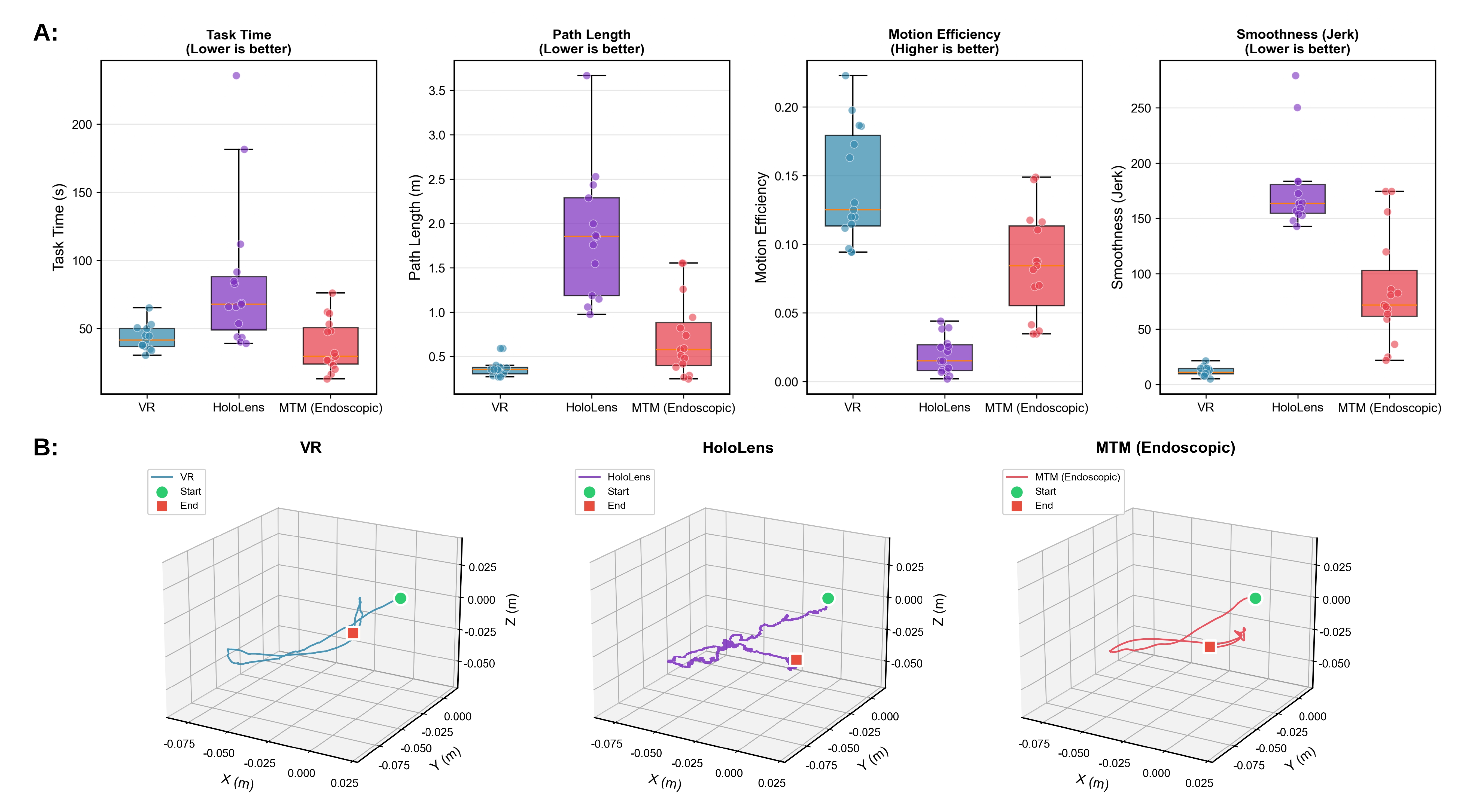}
    \caption{Objective performance metrics and kinematic trajectories across the three teleoperation platforms. (a) Box plots illustrating the distribution of Task Time, Path Length, Motion Efficiency, and Smoothness (Jerk). (b) A sample 3D trajectory comparison from a participant with experience in both dVRK and AR.}
    \label{fig:performance}
\end{figure*}
To evaluate the proposed digital twin-driven VR teleoperation platform, a user study was conducted comparing it against two established paradigms: the traditional stationary console (MTM) of the da Vinci Research Kit (dVRK), and the console-free MR teleoperation system of Magnani et al.~\cite{magnani2026consolefree}, which uses HoloLens~2 with bare-hand tracking and virtual instrument shafts. The objective was to assess how different platforms influence users' depth perception and spatial understanding during a fine manipulation task.

\subsection{Experimental Setup and Protocol}
Participants teleoperated the robotic arm to grasp a vertically oriented 3D-printed ring and place it onto a target peg. Success relied heavily on perceiving tool-to-object depth and accurately orienting the gripper.

The study involved 15 volunteers (9 males, 6 females). Among them, 9 had prior dVRK experience, and 11 had prior VR/AR exposure. To minimize bias, we used a within-subject design where participants tested all systems (MTM, MR, and VR) in a randomized order. After a familiarization phase, participants were instructed to close the gripper only when they believed the jaws were perfectly aligned to grasp the ring. Each trial continued until the participant successfully placed the ring on the peg; therefore, the total number of gripper closures per trial serves as a direct proxy for the operator's confidence in depth estimation, with fewer closures indicating higher spatial certainty. The protocol was approved by the Institutional Review Board (IRB) under protocol HIRB00007467.

\subsection{Evaluation Metrics}
Objective performance was quantified by task completion time and total gripper closures, with fewer closures indicating higher confidence in depth estimation. We evaluated trajectory path length (sum of Euclidean distances between waypoints), motion efficiency (ratio of straight-line
distance to path length) and movement smoothness (quantified by integrated squared jerk~\cite{balasubramanian2015smoothness}). Finally, a custom questionnaire captured subjective feedback on preference, ease of learning, perceived latency, and the specific impact of the multi-view rendering on spatial awareness.

\section{RESULTS AND DISCUSSION}
\label{sec:results}

\subsection{System Latency Evaluation}
\label{subsec:latency}
To quantify the real-time responsiveness of our platform, we independently evaluated the end-to-end latency of (i) the teleoperation control loop that drives the \emph{robot digital twin} and (ii) the perception-driven update loop that drives the \emph{scene digital twin}. Notably, these two latencies are architecturally decoupled: the robot twin is updated from high-rate kinematic state streaming, whereas the scene twin is bounded by the throughput of the perception pipeline.

\subsubsection{Teleoperation Latency}
We estimated the control-loop delay by cross-correlating the time series of target pose commands (from the Quest~3S controllers) with the measured robot poses (returned via UDP), yielding an end-to-end latency of 59.7\,ms.

\subsubsection{Scene Perception Latency and Scalability}
For scene state synchronization, evaluated on an NVIDIA RTX 3080 Ti GPU, we measured the per-frame runtime of the perception modules from execution logs. SAM 2 segmentation required $157.4 \pm 6.5$\,ms, while FoundationPose pose estimation required $50.9 \pm 10.2$\,ms. The scene-twin update rate is primarily compute-bounded and strongly dependent on hardware and inference backend, indicating substantial headroom for acceleration. Official benchmarks report that SAM 2 exceeds 44\,FPS on an NVIDIA A100 GPU, suggesting that the scene perception loop can scale significantly with optimized deployment and scheduling.

\subsection{Task Performance}
\label{subsec:task_performance}

We assessed overall differences using the Friedman test and performed pairwise comparisons with Wilcoxon signed-rank tests (Bonferroni-corrected, $\alpha\!=\!0.05/3$). Table~\ref{tab:stats} summarizes the results across all four objective metrics.

\begin{table}[!t]
\centering
\vspace{1.5mm}
\caption{Performance comparison (mean$\pm$SD) and pairwise Wilcoxon signed-rank $p$-values}
\label{tab:stats}
\setlength{\tabcolsep}{2.5pt}
\renewcommand{\arraystretch}{1.15}
\footnotesize
\begin{tabular}{@{}l c c c c@{}}
\toprule
 & Time (s)$\downarrow$ & Path (m)$\downarrow$ & Effic.$\uparrow$ & Jerk$\downarrow$ \\
\midrule
VR  & 54.6$\pm$32.0 & 0.37$\pm$0.10 & 0.14$\pm$0.04 & 12.0$\pm$4.1 \\
MR  & 85.2$\pm$55.2 & 2.66$\pm$2.21 & 0.02$\pm$0.01 & 245.7$\pm$270 \\
MTM & 37.4$\pm$19.2 & 0.71$\pm$0.44 & 0.08$\pm$0.04 & 86.0$\pm$49.3 \\
\midrule
VR--MR  & .219 & \textbf{$<$.001} & \textbf{$<$.001} & \textbf{$<$.001} \\
VR--MTM & .506 & \textbf{.037} & \textbf{$<$.001} & \textbf{$<$.001} \\
MR--MTM & \textbf{$<$.001} & \textbf{$<$.001} & \textbf{$<$.001} & \textbf{$<$.001} \\
\bottomrule
\end{tabular}
\end{table}

The objective results are visualized in Fig.~\ref{fig:performance}. The MTM console yielded the lowest mean task time (37.4\,s), followed by VR (54.6\,s) and MR (85.2\,s). The VR--MR time difference did not reach pairwise significance after Bonferroni correction ($p\!=\!.219$), largely due to the high variance observed in the MR condition. However, trajectory quality metrics revealed large and statistically robust differences. Compared to the MR baseline, VR reduced path length by 86\% and jerk by 95\% (both $p\!<\!.001$). Most notably, VR also significantly outperformed the MTM console in path length ($p\!=\!.037$), motion efficiency ($p\!<\!.001$), and smoothness ($p\!<\!.001$). This indicates that VR teleoperation produces trajectories that are smoother and more direct than those generated via the physically grounded console.

The high variance and lower performance in the MR condition stemmed primarily from the inherent fragility of bare-hand tracking. Mid-air operation was highly susceptible to hand tremors, tracking loss, and recognition errors, which were further exacerbated by a gesture-based clutching mechanism prone to accidental misoperations. By contrast, the 6-DoF VR controllers eliminated these tracking instabilities, yielding substantially more stable trajectories (Fig.~\ref{fig:performance}b).

Interestingly, despite its physical ergonomic advantages, the MTM console exhibited notably higher jerk than the VR platform ($p\!<\!.001$). This is largely attributable to the absolute orientation alignment inherent in the standard da Vinci clutching mechanism. Upon clutch re-engagement, the master manipulators actively enforce alignment with the current instrument pose. Users who failed to maintain a consistent hand orientation during the clutched repositioning phase experienced abrupt positional transients at re-engagement. This observation highlights a practical drawback of strict physical coupling: the very alignment mechanism designed to provide spatial intuition penalizes users who have not fully internalized the clutch-release dynamics.

Finally, the mean number of gripper closures---a direct proxy for the operator's depth perception confidence---was lowest for VR (1.27), followed by MR (1.47) and MTM (1.73). This suggests that the multi-view VR environment provided the clearest spatial understanding across all conditions.

\subsection{Subjective Evaluation}
\label{subsec:subjective}

\begin{figure}[!t]
    \centering
    \includegraphics[width=0.8\columnwidth]{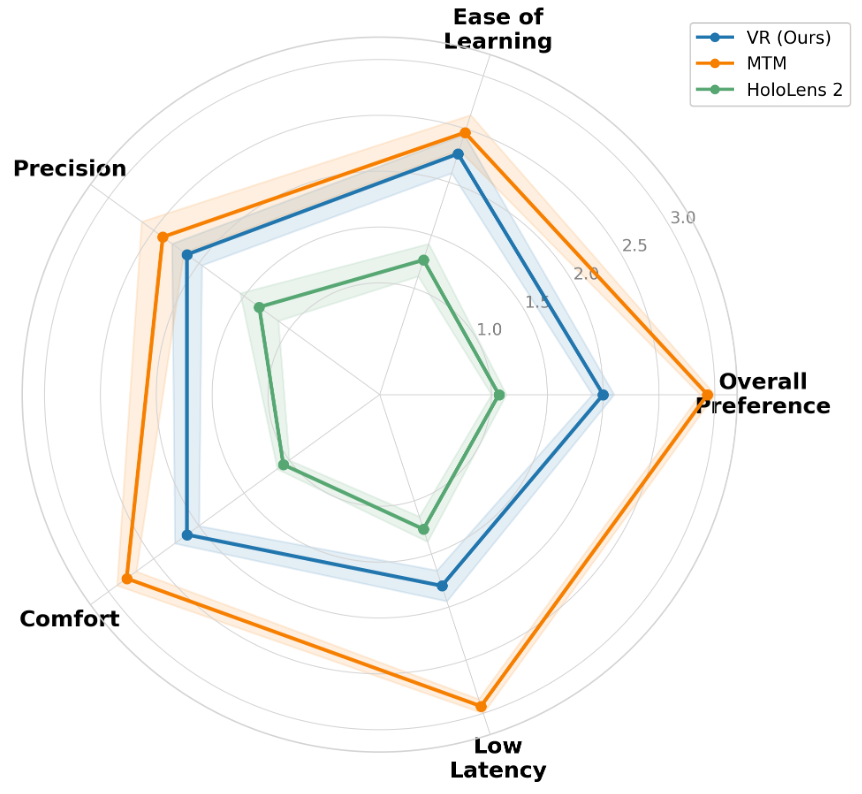}
    \caption{Subjective ranking scores across five dimensions
    (Borda scoring: Rank~1\,=\,3, Rank~2\,=\,2, Rank~3\,=\,1;
    higher is better). Shaded regions indicate $\pm$1 standard error.}
    \label{fig:radar}
\end{figure}

Subjective feedback was quantified using Borda scores across five dimensions (Fig.~\ref{fig:radar}). As expected for a mature, physically grounded interface, the MTM console scored highest overall. However, the proposed VR platform consistently ranked second, substantially outperforming the HoloLens~2 MR baseline. The performance gap between VR and MTM was narrowest in ease of learning (2.27 vs.\ 2.47) and precision (2.13 vs.\ 2.40). Regarding the multi-view rendering feature, 87\% of participants rated it as helpful or very helpful. Among them, 80\% cited improved depth perception and 67\% reported a better understanding of tool pose as the primary benefits. Consequently, 80\% of users preferred the flexible multi-view mode over a fixed endoscopic view.

\subsection{Spatial Perception and Control Decoupling}
\label{subsec:analysis}

Despite the complete absence of physical hand--instrument coupling, VR users produced significantly smoother, more efficient trajectories than MTM users and required the fewest gripper closures across all conditions. Furthermore, 80\% of participants explicitly attributed their improved performance to the enhanced depth perception provided by the multi-view rendering.

These findings support the central observation of this work: the strict reliance on hand--instrument correspondence in existing console-free systems~\cite{magnani2026consolefree} is largely a compensatory response to insufficient spatial perception, rather than a fundamental requirement of teleoperation. When an interface provides sufficiently rich spatial cues, operators can maintain spatial awareness through visual feedback alone. This decoupling safely eliminates the workspace restrictions, visual occlusions, and physical fatigue associated with compensatory virtual-shaft designs (from prolonged posture alignment with the virtual shaft axes without an effective clutching mechanism).

\section{LIMITATIONS AND FUTURE WORK}
\label{sec:limitations}

Several limitations of the current system inform our future work. First, the perception pipeline (FoundationPose) assumes rigid body geometry; extending this real-time digital twin to highly deformable environments, such as soft tissue, remains a significant computational challenge. Second, the system relies on an external RGB-D camera, which is not present in standard clinical setups. Clinical translation will require robust stereo-endoscopic depth estimation algorithms. Third, the scene-twin update rate currently lags behind the 90\,Hz robot-twin rate, though further optimization and hardware acceleration can substantially narrow this gap.

Beyond addressing these limitations, we envision extending this scalable architecture into a versatile practical platform. Having demonstrated its operational superiority in unconstrained teleoperation, the system can naturally be adapted to serve as an auxiliary 3D spatial view within the conventional dVRK endoscopic workflow. Thanks to its rendering from kinematic state rather than high-bandwidth video streams, it can provide surgeons with rich spatial understanding and occlusion resolution alongside the standard camera feed. Furthermore, the platform holds immense potential for scalable data collection. By natively supporting offline replay and synchronized multi-view rendering, it captures kinematic states grounded in real-world physical perception---avoiding the inherent contact inaccuracies of pure simulation. Given that our study already proves that this decoupled VR interface yields superior manipulation stability and trajectory quality, the system can theoretically generate higher-fidelity, expert-level demonstration datasets for deep learning and robot autonomy. We plan to explore these possibilities in our future work.

\section{CONCLUSIONS}
\label{sec:conclusions}

We presented a digital twin-driven VR teleoperation platform for the dVRK that replaces traditional video pass-through with real-time, state-synchronized rendering. Experimental results demonstrate that this approach significantly enhances spatial perception and manipulation stability over an MR baseline, achieving comparable performance to the traditional MTM console. The central finding of this work is that a digital twin-driven platform with rich spatial cues, combined with stable controller input, can collectively offset the need for strict physical hand--instrument correspondence. By safely decoupling the operator's posture from instrument alignment, this paradigm eliminates major ergonomic bottlenecks in existing console-free interfaces.  The underlying state-synchronized architecture further provides a scalable foundation for multi-view data generation and clinical augmentation.

\bibliographystyle{IEEEtran}
\bibliography{references}

\end{document}